\documentclass[runningheads]{llncs}
\usepackage[T1]{fontenc}
\usepackage{graphicx}
\usepackage{hyperref}
\usepackage{color}

\usepackage{algorithm}
\usepackage[algo2e,ruled,linesnumbered]{algorithm2e}
\usepackage{multirow}
\usepackage{booktabs,arydshln}

\usepackage{subfigure}

\usepackage{amsmath}
\usepackage{amssymb}

\usepackage{xcolor}

\begin{document}
\title{One-Shot Pruning for Fast-adapting Pre-trained Models on Devices}
%
%
\author{Haiyan Zhao \and
Guodong Long
}
%
%
\institute{
University of Technology Sydney, Sydney, Australia \\
\email{Haiyan.Zhao-2@student.uts.edu.au} \\
\email{guodong.long@uts.edu.au}
}
\maketitle              
\begin{abstract}

Large-scale pre-trained models have been remarkably successful in resolving downstream tasks. Nonetheless, deploying these models on low-capability devices still requires an effective approach, such as model pruning. However, pruning the model from scratch can pose a practical challenge given the limited resources of each downstream task or device.
To tackle this issue, we present a scalable one-shot pruning method that leverages pruned knowledge of similar tasks to extract a sub-network from the pre-trained model for a new task. Specifically, we create a score mask using the pruned models of similar tasks to identify task-specific filters/nodes in the pre-trained model for the new task. Based on this mask, we conduct a single round of pruning to extract a suitably-sized sub-network that can quickly adapt to the new task with only a few training iterations.
Our experimental analysis demonstrates the effectiveness of the proposed method on the convolutional neural networks (CNNs) and vision transformers (ViT) with various datasets. The proposed method consistently outperforms popular pruning baseline methods in terms of accuracy and efficiency when dealing with diverse downstream tasks with different memory constraints.

\end{abstract}
\section{Introduction}
\label{sec:intro}

Large-scale pre-trained models have exhibited exceptional performance on a wide range of downstream tasks. For instance, CLIP~\cite{radford2021learning} has surpassed the current state-of-the-art computer vision models on 27 downstream tasks, each having diverse distributions. However, these pre-trained models typically consist of millions of parameters, hindering their deployment on edge devices with limited memory and computation budgets.
Previous studies~\cite{xu2022dense,ZHANG202136} have demonstrated that only a subset of the filters/nodes in a pre-trained model are crucial for the inference process of a given downstream task. To address this issue, model pruning presents an effective approach wherein unnecessary filters/nodes can be removed without compromising accuracy.

Conventional pruning methods in real-world applications often require repeated pruning of the pre-trained model to adapt to different downstream tasks and low-capability devices, resulting in a waste of computational power and time. Moreover, some devices may not have the capacity to prune large models from scratch due to memory and computation limitations. The question arises: Is it feasible to find a sparse sub-network within a pre-trained model that can quickly adapt to a new downstream task?
Recent studies~\cite{DBLP:journals/corr/abs-1803-03635,Ye2020GreedyOP,Savarese2020WinningTL} have shown evidence of the lottery ticket hypothesis (LTH), which states that training from a sparse sub-network in a randomly initialized model can achieve comparable performance to the original dense network. However, LTH cannot reduce the number of training iterations required. Furthermore, LTH focuses solely on unstructured weight pruning, which may not necessarily improve the efficiency of training and inference of the pruned model.

\begin{figure*}[htbp]
 \centering
 \includegraphics[width=0.8\columnwidth]{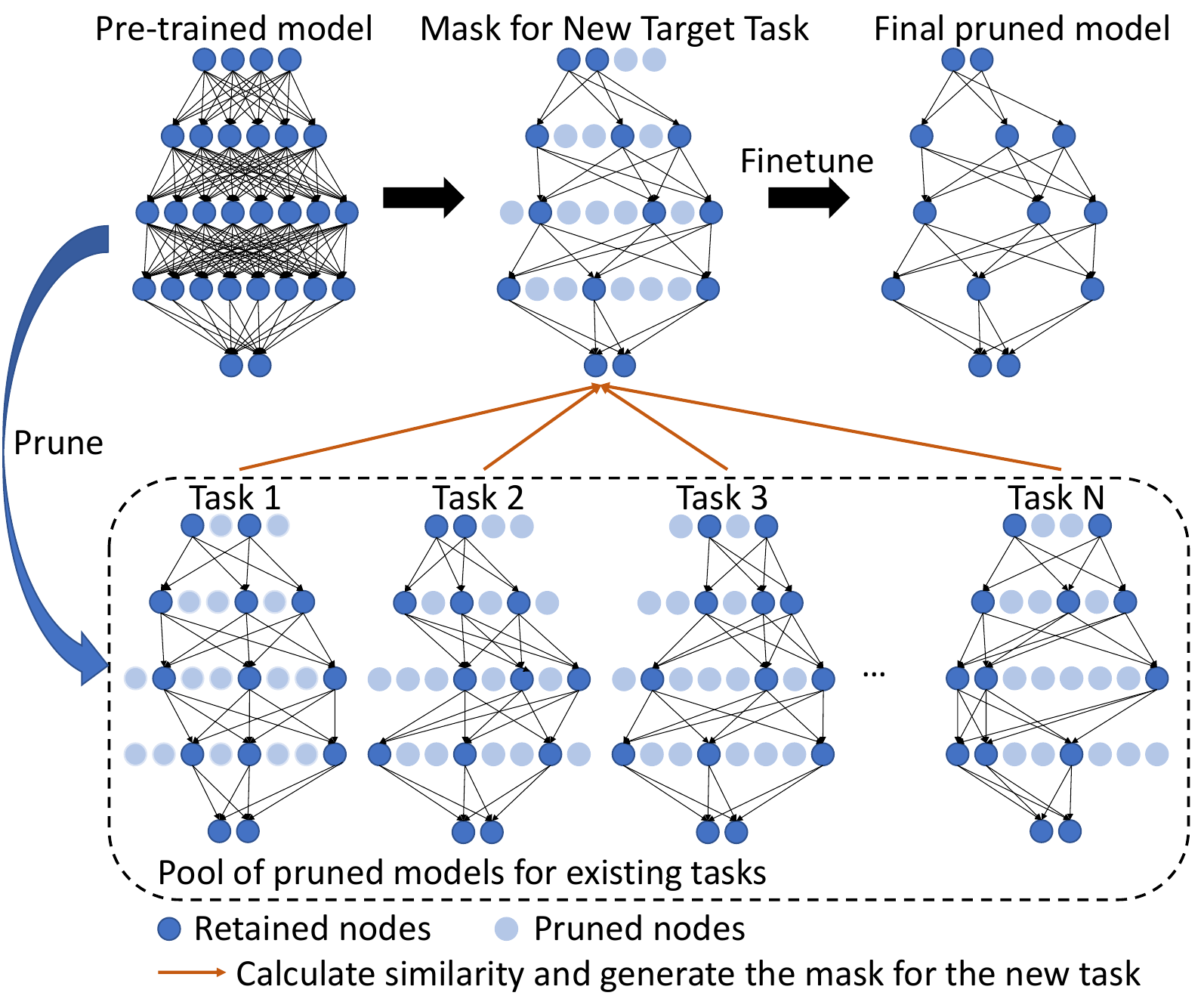}
\caption{
The illustration of the proposed one-shot pruning framework. The framework utilizes the pruned models of existing tasks from a pre-trained model to generate the one-shot pruning mask for each new task. The mask is created by leveraging the similarities between the new task and existing tasks, as shown in Fig.~\ref{fig:smsp}. Once the mask is created, a small number of training iterations are sufficient to obtain the final pruned model.
}
\label{fig:overview}
\end{figure*}

Tian et al.~\cite{tian2020meta} developed a meta-model that is trained on hundreds of tasks to create a well-initialized pruned model, which can rapidly adapt to a new task within a few training iterations, thereby reducing computational costs. The meta-model is the same for all tasks.
However, in practical scenarios, it is common for a pre-trained model to produce pruned models for downstream tasks or devices with varying memory constraints. 
Therefore, we propose to directly utilize prior knowledge from previous pruned models instead of training a new meta-model. For each downstream task, its pruned model retains only critical and task-specific filters/nodes from the pre-trained model. 
We investigate the relationship between the pruned models of downstream tasks with different similarities. We observe that tasks with high similarities share more task-specific filters/nodes in their pruned models.

Based on this observation, this paper proposes a novel one-time pruning method called "Scalable Mask Selection Pruning (SMSP)", which is illustrated in Fig.~\ref{fig:overview}. By learning from the pruned results of similar tasks, SMSP can create a mask to identify task-specific filters/nodes in the pre-trained model and prune the model once to extract a suitably sized sparse sub-network for a new task. SMSP is scalable because the created mask can be used to extract a sub-network of any pruning ratio from the pre-trained model to adapt to different devices. The sparse sub-network is then trained on the training data of the new task for a few iterations to quickly adapt to the new task. SMSP can significantly reduce the computation cost during pruning while maintaining the excellent performance of the pruned models. Extensive experiments have been conducted to evaluate the proposed method, which demonstrates that SMSP outperforms state-of-the-art pruning methods on CNN and ViT over several datasets. Furthermore, SMSP performs well when used to produce pruned models for tasks with different memory constraints and tasks from unseen datasets, which demonstrates the scalability and generality of SMSP.

\section{Related Works}

Model pruning is a highly effective technique for compressing deep neural networks.
Some existing works~\cite{Han2015LearningBW,Li2017PruningFF,he2019filter} apply iterative pruning approaches to reduce the model size by eliminating filters/nodes with small weights while minimizing the loss of accuracy.
Alternatively, methods like HRank~\cite{lin2020hrank} and APoZ~\cite{hu2016network} evaluate the importance of each filter based on its corresponding activation maps.
Another line of methods~\cite{lin2018accelerating,lin2020dynamic} maintains a mask for filters/nodes in the model to eliminate redundant parameters automatically. 
And this dynamic pruning setting is also widely used in the pruning of the vision transformer. 
Recent works~\cite{yu2022unified,yu2022width,zhu2021vision} introduce learnable parameters to each attention head, node, layer, or block in the vision transformer to reduce the model's complexity.
The approach of Goyal et al.~\cite{goyal2020power} is different from traditional parameter pruning as they dynamically prune input patches in each block of ViT, resulting in significant reductions in inference computation without compromising the model's performance. Meanwhile, Tang et al.\cite{tang2022patch} evaluate the importance of each patch in maintaining the original final results. However, these pruning methods require starting the pruning process from scratch, which is time-consuming. In contrast, our method leverages pruned knowledge of similar tasks to reduce the number of pruning iterations significantly.

Some other pruning methods aim to speed up the pruning process. 
Cai et al. \cite{cai2020once} propose a once-for-all network that supports diverse settings by decoupling training and neural architecture search, which reduces the cost and makes it scalable for efficient inference across many devices and resource constraints. However, the generated pruned models are all for one task and cannot be generalized to other tasks.
Tian et al.\cite{tian2020meta} proposed a meta method that trains a well-initialized pruned meta-model to quickly adapt to different few-shot tasks. However, this meta-model is the same for all tasks and cannot generalize to devices with varying memory constraints.
MEST\cite{yuan2021mest}, which is designed for edge devices, starts training from a sparse sub-network to save training computation.
DLTH~\cite{bai2022dual} is a variant of LTH and also starts from a well-designed sub-network. It claims that randomly extracted subnetworks from a randomly initialized dense network can be transformed into a well-performing sub-network that can achieve admirable performance compared to LTH. However, all these methods require a significant amount of time and computation to find the initialized sub-networks. In contrast, our proposed method can be applied to different downstream tasks, and it does not require any additional computation cost to extract a sub-network for each new task.

\section{Methodology}

In this section, we establish a model pool consisting of the pruned models obtained from hundreds of tasks on both CNN and ViT. These pruned models are extracted to retain the task-specific knowledge present in the pre-trained model for each task. 
We observe that similar tasks tend to share more task-specific filters/nodes. Leveraging this observation, we propose a generic and scalable approach to reduce the computational cost of pruning for new tasks or devices.

\subsection{Pool of Pruned Models from Different Tasks}
\label{sec:modelzoo}

A pruned model of the downstream task typically preserves filters/nodes that are indispensable for its inference in the pre-trained model. 
In practice, a dataset of pruned models exists owing to the extensive utilization of large-scale models across various downstream tasks and devices.
In this paper, to emulate this situation, we construct a simplified dataset of pruned models for different tasks and devices using the same pre-trained models.

\begin{algorithm}[ht]
    \SetKwInOut{Input}{Input}
    \SetKwInOut{Output}{Output}
    \SetKwInOut{Init}{Initialize}
    \Input{Pre-trained network $F(\cdot; \Theta)$, Task $t$ and training set $D_t$, training iterations $J$, target pruning ratio $r$, pruning threshold $\tau$.}
    \Init{$\Omega\leftarrow[n]$, the set of filters/heads/nodes preserved in the network. $S^{t}_{i}\leftarrow1$, the mask score for each prunable filter/head/node for task $t$.}
    \For{$j \gets 1$ \textbf{to} $J$}{
            \For{$i\in\Omega$}{
                Prune the filter/node $\theta^{t}_{i}$ from $\Omega$ if its score $S^{t}_{i}<\tau$, $S^{t}_{i}\leftarrow0$;
                }
        Stop pruning if the pruning ratio $1-(|\Omega|/n)$ $\ge$ $r$;
        
        Apply one optimization step on a mini-batch of $D_t$ according to Eq.~(\ref{equ:amploss}) to fine-tune the remained filters/heads/nodes and mask scores $\{\theta^{t}_{i},S^{t}_{i}: i\in\Omega\}$.}

\caption{\sc{Automatic Mask Pruning}}
\label{alg:automatic_pruning}
\end{algorithm}

\textbf{Automatic Mask Pruning (AMP).} 
Inspired by \cite{liu2017learning,yu2022unified}, we propose automatic mask pruning (AMP) to automatically identify task-specific filters/nodes for different tasks in the pre-trained model. 
Algorithm~\ref{alg:automatic_pruning} provides a detailed outline of the AMP process.
Specifically, given a pre-trained network $F(\cdot;\Theta)$ with parameter $\Theta$ and a training set $D_t$ of a new target task $t$, let $\Theta^t=\{\theta^{t}_{i}\}_{i=1:n}$ where $\theta^{t}_{i}$ denotes every filter/head/node-$i$ in the network.
By adding a mask, we incorporate a learnable mask score $S^{t}_{i}$ to each prunable filter/head/node-$i$ in the pre-trained model.
We define an operator $\odot$ applied to $\Theta^{t}$ and its associated scores $S^{t}$ as
\begin{small}
\begin{equation}
(\Theta^{t} \odot S^{t})[i]\triangleq \Theta^{t}[i] \cdot S^{t}[i]
\end{equation}
\end{small}

During the pruning process, these differentiable scores are optimized along with model parameters. To encourage sparsity, an additional L1 regularization loss is applied and filters/nodes with scores below a predefined threshold will be pruned.
The final objective function of AMP is defined as follows:
\begin{small}
\begin{equation}
\label{equ:amploss}
\min\limits_{\{S^t_i\}_{i=1:n}} \mathbb{E}_{(x,y)\sim D_t}l(y, F(x;\Theta^t\odot S^t))+ \lambda\|S^t\|_1
\end{equation}
\end{small}
where $y$ represents the ground truth for $x$, $l$ denotes the cross-entropy loss, and $\lambda$ is the weight used to balance between the two losses.

We apply AMP to prune two major categories of pre-trained models, i.e., CNN and ViT, for diverse tasks with different memory constraints.
Specifically, we select ResNet-18(ResNet-50)\cite{He2016DeepRL} pre-trained on CIFAR-100\cite{CIFAR}(ImageNet~\cite{ImageNet}) for CNN, and apply AMP to multiply the mask score to each filter in the network.
For ViT, we use DeiT-S~\cite{touvron2021training} pre-trained on ImageNet.
As reported in previous work~\cite{ZHANG202136}, only some attention heads in deep pre-trained transformers are necessary for downstream tasks. Therefore, AMP is used to prune ViT at two levels: heads in the multi-head attention modules and nodes in the feed-forward network modules.

In the pool of pruned models, tasks for ResNet-18, ResNet-50, and ViT are randomly sampled from classes in CIFAR-100 and ImageNet datasets, respectively.
To verify the generality and scalability of our proposed method, we collect the pruned models of diverse tasks,
which can be divided into $3$ groups: $3$-classes, $5$-classes and $10$-classes classification tasks, each containing $300$ tasks.
To emulate the memory limitations of various devices, we store pruned models with varying pruning ratios for each task in our model pool.
Due to the high memory costs of storing each pruned model, we have modified the AMP algorithm such that only mask scores are optimized with regularization, while all pre-trained model parameters remain fixed.
This modification facilitates accurate masking for each task to identify task-specific knowledge in the pre-trained model. 
As all tasks can share the same pre-trained model during inference, we only record the class labels $C^t$ and the mask $S^t$ for each task $t$. The mask scores of pruned filters/nodes are then set to $0$.

\subsection{Knowledge Shared between Tasks}
\label{sec:emp_study}

In the realm of multi-task/lifelong learning methods, similar tasks usually share more parameters in the network. In this section, we study the overlap of pruned models for similar tasks to verify whether more similar downstream tasks share more parameters in the pre-trained model.
To compute the similarity between downstream tasks, we apply the Log Expected Empirical Prediction (LEEP)~\cite{nguyen2020leep}, which is used to evaluate the transferability of representations learned by the source task to the target task. This method only requires running the target task's data through the pruned model once to compute the LEEP score.

\begin{figure*}[htbp]
\vspace{-1em}
     \centering
     \subfigure[Overlap of Filters for ResNet-18]{
         \centering
         \includegraphics[width=0.42\columnwidth]{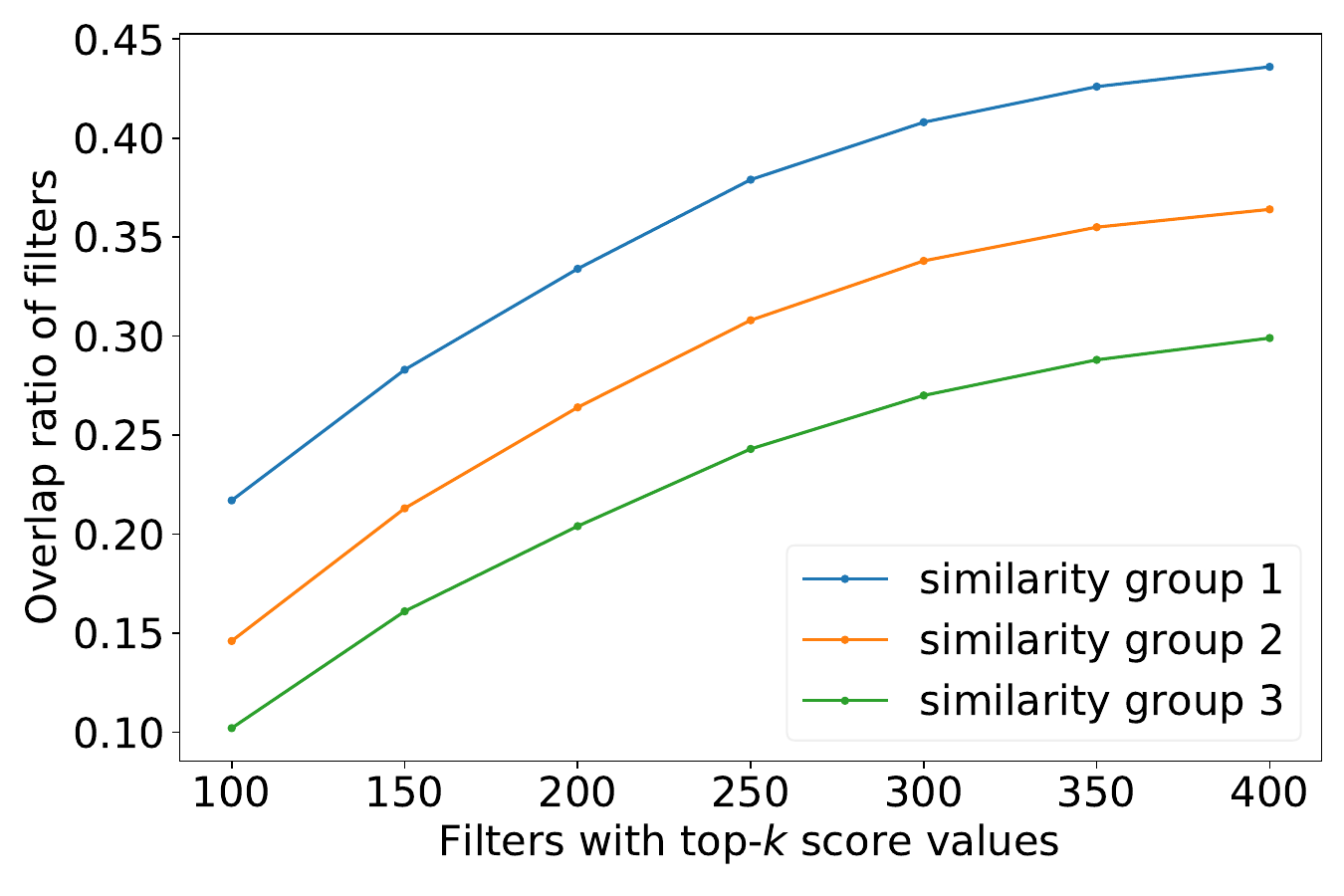}
     }
     \subfigure[Overlap of Filters for ResNet-50]{
         \centering
         \includegraphics[width=0.42\columnwidth]{img/imagenet_nodes_overlap.pdf}
     }
     \subfigure[Overlap of Heads for ViT]{
         \centering
         \includegraphics[width=0.42\columnwidth]{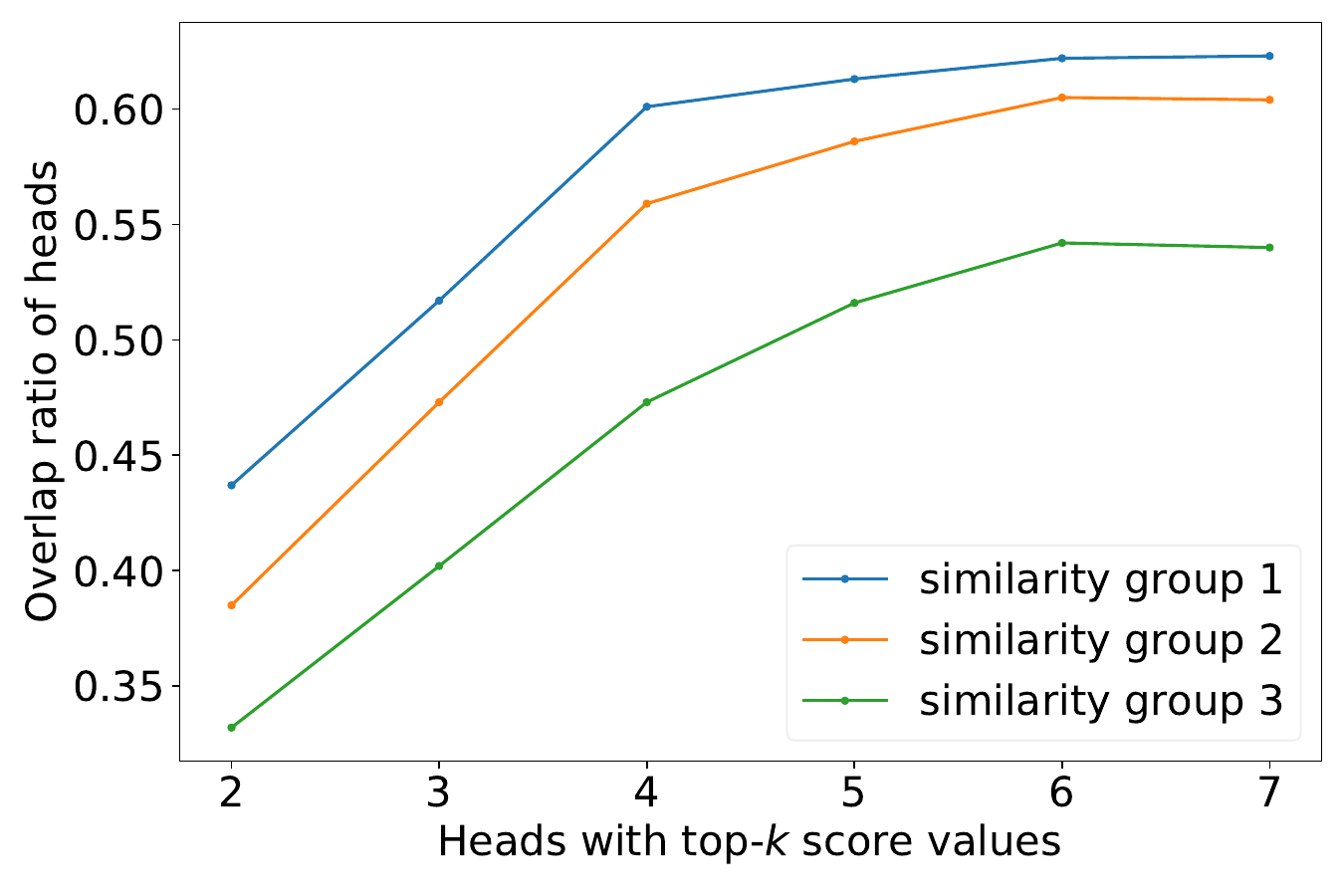}
     }
    \subfigure[Overlap of Nodes for ViT]{
             \centering
             \includegraphics[width=0.42\columnwidth]{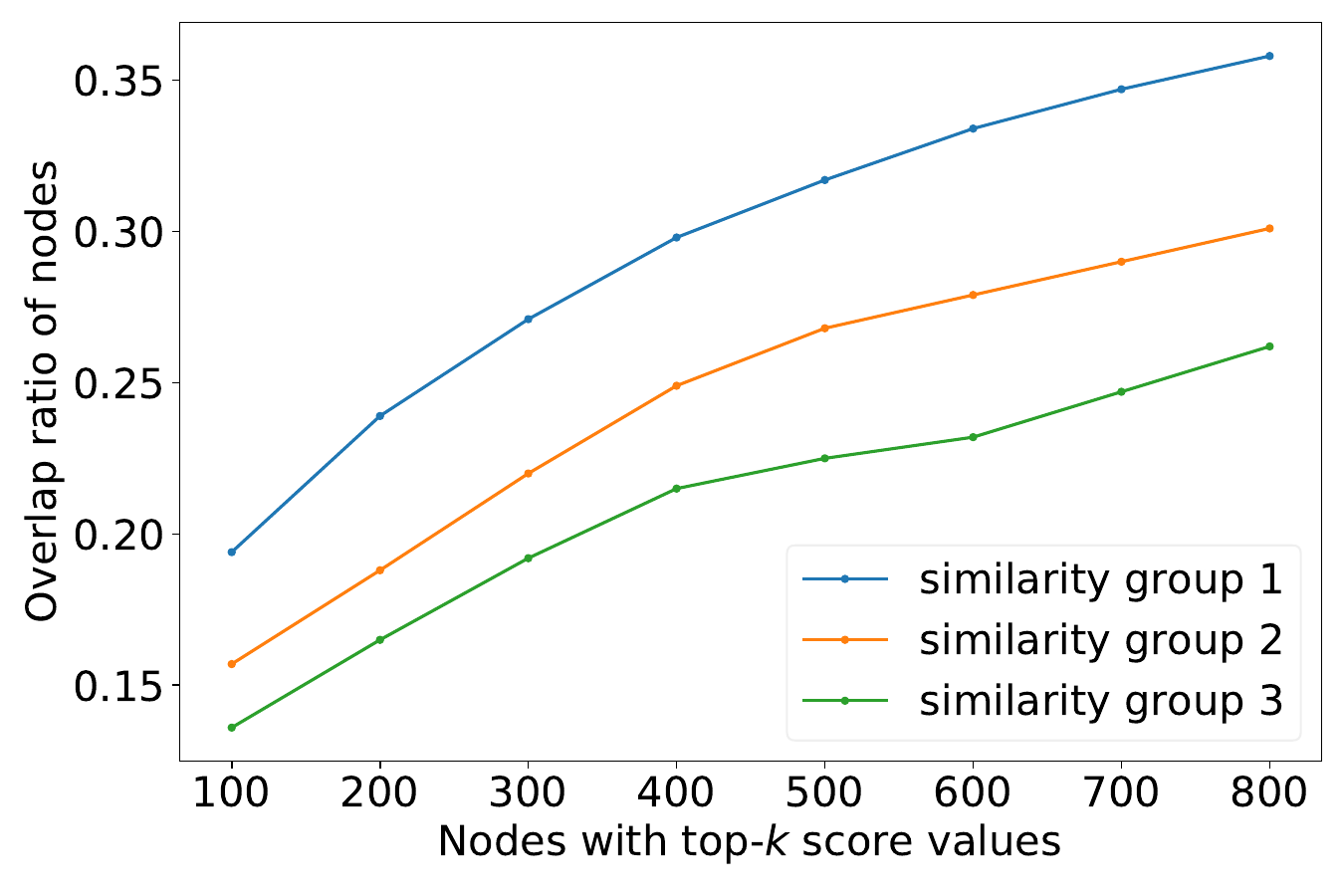}
         }
\vspace{-1em}
\caption{
The overlap ratios of task-specific filters/heads/nodes retained by the tasks with different similarities in ResNet-18 or ViT.
 The x-axis denotes the top-$k$ filters/heads/nodes with the highest mask scores in the pruned model. 
 From similarity groups 1 to 3, the similarities between tasks within each group decrease. \textbf{Tasks with higher similarities tend to share more task-specific knowledge.}
}
\vspace{-1em}
\label{fig:overlap_ratio}

\end{figure*}

\textbf{Overlap of task-specific filters/nodes.}
Upon applying AMP to a new task, filters or nodes that have small mask scores will be pruned, whereas those with high mask scores, which contain task-specific knowledge relevant to the downstream task, can be retained in the model.
So we focus on the overlap of these high-score filters/nodes between tasks.
Given the pruned model of a task $m$, the set of filters/nodes $\Omega^m$ retained in the pre-trained model are sorted according to their mask scores $\{S^{m}_{i}\}_{i\in\Omega^m}$ in the descending order.
$\Omega^m_k$ denotes the filters/nodes with top-$k$ mask score values in the mask of task $m$.
For each pair of tasks, say task $m$ and task $n$ (using the same pre-trained model), we compute the overlap ratio $R$ of filters/nodes with top-$k$ score values in their masks, i.e., $R = |\Omega^m_k \cap \Omega^n_k|/k$.

In Fig.~\ref{fig:overlap_ratio}, we present the overlap ratio of retained filters/nodes in various pre-trained models for tasks with varying degrees of similarity.
The x-axis of Fig.~\ref{fig:overlap_ratio} represents the top-$k$ filters/heads/nodes with the highest mask scores in the pruned model, while the y-axis represents the overlap ratio of top-$k$ filters in the pruned models of two similar tasks.
Given a new task, we calculate its LEEP similarities to the existing tasks in the model pool. Then we sort these LEEP similarities and partition them into three groups of equal intervals. Existing tasks whose similarity scores fall into a specific interval will be assigned to the corresponding similarity group. From similarity group 1 to group 3 in Fig.~\ref{fig:overlap_ratio}, the similarities between tasks decrease. 

\begin{figure*}[htbp]
 \centering
 \includegraphics[width=0.8\columnwidth]{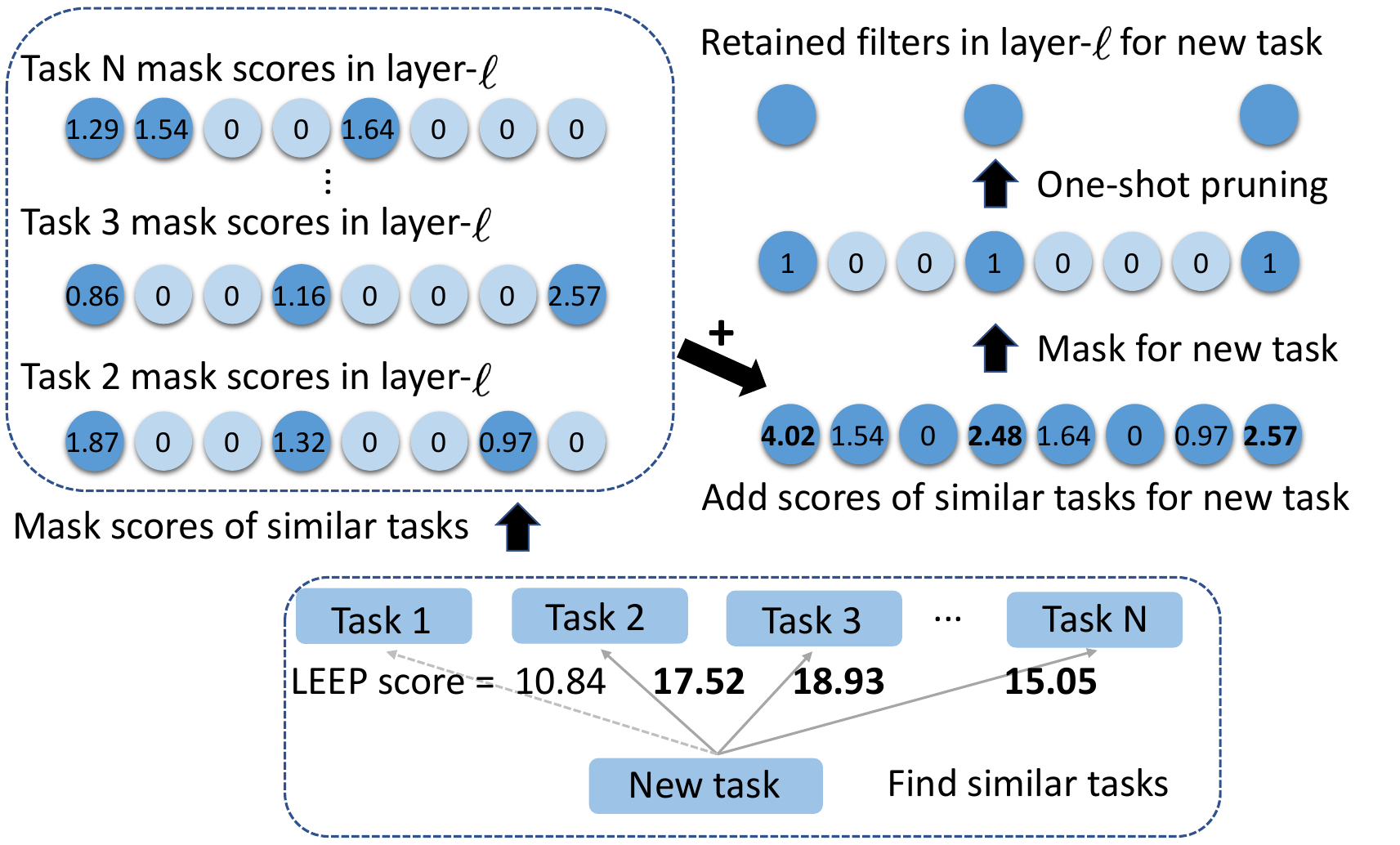}
\caption{
An example of generating the mask for a new task. The mask scores of each similar task are summed up to generate mask scores for the new task. Then, filters/nodes with high scores in layer-$\ell$ will be preserved by the new task.
}
\label{fig:smsp}
\end{figure*}

We observed from all three plots in Fig.~\ref{fig:overlap_ratio} that the overlap ratios of tasks belonging to similarity group 1 are considerably greater than those of tasks in similarity group 3. This indicates that the pruned models of \textbf{more similar tasks share a significantly higher number of task-specific filters/heads/nodes}. Hence, the pruned models of previous similar tasks can be utilized to identify task-specific parameters in the pre-trained model, expediting the pruning of the new task.
On the other hand, as the value of $k$ increases, the overlap ratios in three plots grow gradually. This can be attributed to the fact that certain filters/heads/nodes with high mask scores in one task may be retained by another task with smaller scores. These filters/nodes have varying importance for different tasks and may serve distinct roles. In plot (c), we observe that the overlap ratios begin to converge when $k$ exceeds $6$. This is due to the fact that only a small number of heads (approximately 8) are preserved in the pruned model of each task.

\subsection{Scalable Mask Selection Pruning (SMSP)}
\label{sec:SMSP}

Inspired by the above discovery, we propose a generic and simple method called ``Scalable Mask Selection Pruning (SMSP)" to fast-adapt the pre-trained model to downstream tasks.
The process of generating a mask for each new task is illustrated in Figure~\ref{fig:smsp}.
SMSP leverages the knowledge of pruned models for similar tasks to create a pruning mask of the pre-trained model for a new task. The detailed process of SMSP is shown in Alg.~\ref{alg:scale-pruning}.

Specifically, given a new task $t$, SMSP first calculates its LEEP similarities~\cite{nguyen2020leep} to tasks in the pool and samples $M$ similar neighbor tasks $M^t$. 
The mask scores $S^t$ of task $t$ are computed by summing the mask scores of all selected similar tasks, as shown below:
\begin{small}
\begin{equation}
\{S^t_i\}_{i=1:n} = \sum_{m=1}^MS^m_i
\label{equ:score_sum}
\end{equation}
\end{small}
Here, $n$ represents the total number of filters/heads/nodes in the model, and $M$ represents the total number of selected similar tasks.

\begin{algorithm}[htbp]
    \setcounter{AlgoLine}{0}
    \SetKwInOut{Input}{Input}
    \SetKwInOut{Output}{Output}
    \SetKwInOut{Init}{Initialize}
    \Input{New task $t$ and its training set $D_t$, pruning ratio $r$, training iterations $J$, number of similar tasks $M$, a pool of pruned models.}
    \Output{A pruned model for new task-$t$}
    \Init{$\Omega\leftarrow[n]$, the set of filters preserved by pruned model of task $t$, $\{S^t_i\leftarrow0\}_{i=1:n}$, the mask scores of prunable filters/nodes for task $t$}
    Sample/find $M$ similar neighbour tasks $M^t$ to task $t$ according to LEEP similarity;\par
    Calculate $\{S^t_i\}_{i=1:n}$ according to Eq.~(\ref{equ:score_sum}), and sort them;
    \par
    Calculate the number of pruned filters/nodes $k=n*r$;\par

    $s\leftarrow$ the $k$-th smallest score in $\{S^t_i\}_{i=1:n}$;\par

    \For{$i \gets 1$ \textbf{to} $n$}{
            Prune filter/node $i$ from $\Omega$ if $S^t_i<s$;
           }
    Train $\theta^{t}_{i}: i\in\Omega$ for $J$ iterations on $D_t$.\par
\caption{\sc{Scalable Mask Selection Pruning}}
\label{alg:scale-pruning}
\end{algorithm}

As filters/nodes with high scores in $S^t$ have been shown to play essential roles in similar tasks, it is likely that they contain task-specific knowledge relevant to the new target task $t$.
We sort the mask score of task $t$ in descending order. Given any pruning ratio $r$, SMSP prunes $r*n$ filters with the smallest mask scores once to meet the memory constraint.
The training objective of SMSP is:
\begin{small}
\begin{equation}
\label{equ:smsploss}
\min~ \mathbb{E}_{(x,y)\sim D_t}l(y, F(x;\theta^{t}_{i}: i\in\Omega))
\end{equation}
\end{small}
where $\theta^{t}_{i}: i\in\Omega$ represents filters/nodes retained after pruning.
In the retained sub-network, the mask is removed, and all the parameters are inherited from the original pre-trained model.
SMSP trains the sub-network on the new target task's data for only a few iterations to speed up pruning.

\section{Experiments}

In this section, we evaluate SMSP by pruning ResNet and ViT for downstream tasks from several datasets and compare its results with SOTA pruning methods. We validate the scalability and generality of SMSP by generating pruned models for tasks with different memory constraints. Finally, we study the effect of the mask, the number of similar tasks and task similarities on SMSP.

\subsection{Experimental Settings}

For each experiment scenario, we randomly sample $50$ test tasks from the dataset. Each test task selects its similar tasks from the pool of pruned models according to their LEEP similarities. To make our study more solid, classes in selected similar tasks are disjoint from those in the test task so that their training data are totally different.
In our experiments, we conduct a grid search on a small subset of test tasks to tune the hyperparameters, which are then applied to all tasks. When applying SMSP to prune ResNet, we utilize SGD to train the sub-network and apply cosine annealing learning rate. The batch size is set to 128, and the initial learning rate is set to 0.01.
For experiments of ViT, we follow previous works\cite{touvron2021training} and use the optimizer of AdamW with the cosine-annealing learning rate. During training, we use a batch size of 256 and a smaller initial learning rate of 0.0002.
All results shown in this section are averaged over $50$ test tasks.

\begin{table}[htbp]
\begin{center}
\caption{ Comparison between SMSP with baselines for CNN
}
\resizebox{\columnwidth}{!}{
	\begin{tabular}{lcccccc}
		\toprule
		\multirow{2}*{Methods}  & \multicolumn{3}{c}{\textbf{ResNet-18}} & \multicolumn{3}{c}{\textbf{ResNet-50}} \\
		  & Acc(\%) & FLOPs(T) & Iters & Acc(\%) & FLOPs(T) & Iters  \\
		\midrule
AMP                                            & 87.63$\pm$0.89 & 13.05 & 1000 & 87.12$\pm$0.83 & 108.68 & 1000 \\
Feature Pruning~\cite{molchanov2016pruning}    &87.63$\pm$0.52 & 14.88 & 1000 & 90.89$\pm$0.75 & 110.06 & 1000  \\
Taylor Pruning~\cite{molchanov2019importance}  & 88.32$\pm$0.63 & 14.88 &  1000 & 91.23$\pm$1.17 & 110.06 & 1000 \\
IHT-based Reptile~\cite{tian2020meta}          & 75.73$\pm$0.64 & 0.43  & 100  & 74.29$\pm$0.87 & 3.16 & 100\\
DLTH~\cite{bai2022dual}                        & 74.88$\pm$0.92 & 4.28  & 100  & 70.31$\pm$2.63 &  31.64 & 100 \\
MEST~\cite{yuan2021mest}                       & 77.15$\pm$0.76 & 0.47  & 100  & 68.93$\pm$3.08 & 3.48 & 100 \\
SMSP(\textbf{ours})                            & 88.55$\pm$0.26 & 0.43 & 100 &  90.65$\pm$0.52 & 3.16 & 100\\
		\bottomrule
	\end{tabular} }
	\label{tab:resnet_results}
\end{center}
\end{table}

\subsection{Comparison with SOTA Methods}

We compare our method with several SOTA pruning methods. To demonstrate our method's effectiveness, we compare it with AMP, a conventional pruning method that prunes the pre-trained model from scratch using a large number of pruning iterations.
For tasks on ResNet, we also include two popular pruning methods as baselines: Feature Pruning~\cite{molchanov2016pruning} and Taylor Pruning~\cite{molchanov2019importance}. Feature Pruning calculates the importance of filters by averaging the activation values over all training samples, while Taylor Pruning measures the impact of removing each filter on the final loss to determine their importance.
We also compare our method with some popular methods that accelerate pruning. For example, IHT-based Reptile~\cite{tian2020meta} learns a well-initialized pruned meta-model on a set of training tasks. For each new task, it can obtain the final pruned model by training the meta-model for a few iterations. DLTH~\cite{bai2022dual} is a variant of LTH, which extracts a winning ticket for each task. MEST~\cite{yuan2021mest} can accelerate pruning by training from a sparse sub-network.
For pruning ViT, we compare SMSP with PoWER~\cite{goyal2020power}, which proposes to dynamically prune the input patches of each block in ViT, and UVC~\cite{yu2022unified}, which not only prunes heads and nodes but also unimportant layers and blocks in the model.

\begin{table}[ht]
\small
\begin{center}
\caption{ Comparison between SMSP with baselines for ViT.}
\resizebox{\columnwidth}{!}{
\begin{tabular}{lccc}
    \toprule
    Methods & Accuracy(\%) & FLOPs(T) & Training Iterations \\
    \midrule
    AMP                           & 89.83$\pm$0.46 & 81.50 & 1000  \\
    UVC~\cite{yu2022unified}      & 81.32$\pm$0.87 & 26.73 & 100 \\
    PoWER~\cite{goyal2020power}   & 78.61$\pm$1.32 & 20.86 & 100  \\
    SMSP(\textbf{ours})                    & 90.24$\pm$0.35 & 3.25 & 100 \\
    \bottomrule
\end{tabular} }
\label{tab:vit_results}
\end{center}
\end{table}

The results of comparing SMSP with the baseline methods for ResNet and ViT are presented in Tab.~\ref{tab:resnet_results} and Tab.~\ref{tab:vit_results}, respectively. All results are obtained by pruning $5$-classes classification tasks with a pruning ratio of $90\%$. The findings indicate that, for both ResNet and ViT, SMSP performs slightly better than AMP, which requires significantly more pruning iterations. 

Although Feature Pruning and Taylor Pruning also yield similar or slightly better results than SMSP for ResNet-18 and ResNet-50, they demand significantly more computational resources than SMSP.
Moreover, SMSP surpasses IHT-based Reptile by a large margin, despite the fact that both approaches leverage knowledge from multiple tasks. Unlike IHT-based Reptile, which employs the same pruned meta-model for each new task, SMSP extracts different sub-networks for different tasks, composed of task-specific parameters, which can enhance performance.
Furthermore, the performance of SMSP outperforms DLTH and MEST, which, like SMSP, start with a well-designed sub-network. However, neither DLTH nor MEST has task-specific knowledge in their initialized pruned model, while SMSP initializes the sub-network by leveraging knowledge from similar tasks.

The outcomes presented in Tab.~\ref{tab:vit_results} demonstrate that SMSP significantly outperforms baseline methods for ViT. Owing to a relatively low number of training iterations, neither UVC nor PoWER can recover the accuracy when a considerable number of parameters or patches are eliminated. Conversely, SMSP leverages a sub-network created by similar tasks as an initialization, hence, only a few training iterations are necessary to construct a well-performing pruned model.

\subsection{Evaluation of Scalability and Generality}

Our proposed SMSP is scalable in two folds. 1) SMSP can produce a promising pruned model for a new task of any memory constraint with a few training iterations. 2) All pruned models of tasks with varying data distribution and sizes can be selected as similar tasks to accelerate the pruning of the new task.

\begin{table}[ht]
\small
\begin{center}
\caption{Results of SMSP on tasks of different sizes.}
\resizebox{\columnwidth}{!}{
\begin{tabular}{lcc}
    \toprule
    & 3-classes test tasks & 5-classes test tasks  \\
    \midrule
    3-classes similar tasks      & 93.53$\pm$0.14 & 87.69$\pm$0.22 \\
    5-classes similar tasks      & 93.99$\pm$0.18 & 88.55$\pm$0.26 \\
    \bottomrule
\end{tabular} }
\label{tab:resnet_tasks_size}
\end{center}
\end{table}

\textbf{Applying SMSP to tasks of different sizes.} 
In Tab.~\ref{tab:resnet_tasks_size}, we show the results of applying SMSP to  tasks of different sizes. The pruning ratios of all tasks are set to $90\%$. In the table, we find that for test tasks of different sizes, when we use the 5-classes similar tasks to extract the sub-networks for the test tasks, its performance is better than that of the 3-classes similar tasks. This is because similar tasks containing more classes can better differentiate data from different classes. Similar tasks of large sizes can extract more accurate task-specific filters/nodes for a given new task.

\textbf{Applying SMSP to tasks of different memory constraints.}
In Tab.~\ref{tab:resnet_tasks_memory}, we apply SMSP to tasks of varying memory constraints. All the tasks are 5-classes classification tasks. 
We observe that SMSP outperforms AMP when transferring between different pruning ratios. 
Additionally, SMSP performs better when the pruning ratios of similar tasks and test tasks are the same. This could be attributed to the fact that in a pruned model with a small pruning ratio, some redundant filters/nodes are preserved in the mask, whereas in a pruned model with a large pruning ratio, some task-specific filters/nodes will be removed.
An interesting finding is that SMSP can leverage similar tasks with large pruning ratios to generate a well-performing pruned model of a smaller pruning ratio for a new task. This demonstrates the superiority of using pruned results of similar tasks as prior knowledge.

\begin{table}[ht]
\small
\begin{center}
\caption{ Results of SMSP on tasks of varying memory limits, pr means pruning ratio.}
\resizebox{\columnwidth}{!}{
\begin{tabular}{lccccc}
    \toprule
    & 85\% pr test tasks & 87\% pr test tasks & 90\% pr test tasks & 92\% pr test tasks & 95\% pr test tasks  \\
    \midrule
    AMP                        & 87.14$\pm$0.70 & 86.22$\pm$0.55 & 87.63$\pm$0.89 & 81.02$\pm$0.85 & 71.70$\pm$1.15 \\
    85\% pr similar tasks      & 91.72$\pm$0.17 & 90.52$\pm$0.37 & 87.35$\pm$0.31 & 83.42$\pm$1.05 & 71.56$\pm$1.11 \\
    90\% pr similar tasks      & 90.88$\pm$0.24 & 90.41$\pm$0.43 & 88.55$\pm$0.26 & 85.19$\pm$0.78 & 73.62$\pm$1.36 \\
    \bottomrule
\end{tabular} }
\vspace{-2em}
\label{tab:resnet_tasks_memory}
\end{center}
\end{table}


\textbf{Performance on unseen tasks.} To validate the generality of SMSP, we randomly sample $50$ test tasks from Caltech-256~\cite{griffin_holub_perona_2022}. SMSP produces pruned models for these test tasks by learning from pruned results of tasks from ViT trained on ImageNet. The pre-trained ViT and similar tasks in the pool of pruned results never see the data of Caltech-256. All the test tasks are 5-classes classification tasks with the pruning ratio of $90\%$. 
In Tab.~\ref{tab:caltech_results}, we show the results of applying SMSP to Caltech-256 and compare it with AMP. 
The results show that SMSP can achieve comparable performance as AMP, which uses 10x training iterations. This indicates that SMSP can also identify task-specific heads/nodes in the pre-trained ViT for each unseen task from Caltech-256, so only a few training iterations suffice to produce a well-performed pruned model, showing the generality of SMSP to diverse datasets.

\begin{table}[ht]
\vspace{-0.5em}
\small
\begin{center}
\caption{ Accuracy of applying SMSP to unseen tasks.}
\resizebox{\columnwidth}{!}{
\begin{tabular}{lccc}
    \toprule
    Methods & Accuracy(\%) & FLOPs(T) & Training Iterations \\
    \midrule
    AMP                           & 81.53$\pm$0.92 & 29.82 & 300  \\
    SMSP(\textbf{ours})           & 80.64$\pm$0.69 & 1.96  & 60 \\
    \bottomrule
\end{tabular} }
\vspace{-2em}
\label{tab:caltech_results}
\end{center}
\end{table}

\subsection{Ablation Study}

\textbf{Effect of the mask.}
The main contribution of SMSP is its ability to leverage the pruned results of similar tasks to generate the task-specific mask for each new test task. To validate the efficacy of the masks produced by SMSP, we randomly generate a mask for each task using the same pruning ratio and compare their performance with that of SMSP. In Tab.~\ref{tab:random_mask}, we observe that for tasks using ResNet-18 and ViT, the performance of random masks is significantly worse than that of SMSP. These results suggest that the masks generated by SMSP can effectively identify filters/nodes that are relevant to the new target tasks.

\begin{table}[ht]
\small
\begin{center}
\caption{ Comparison of accuracy between different masks.}
\resizebox{\columnwidth}{!}{
\begin{tabular}{lcc}
    \toprule
    Methods & Accuracy(\%) of ResNet-18 &  Accuracy(\%) of ViT \\
    \midrule
    Random Masks    & 37.78$\pm$8.16 & 60.62$\pm$6.98  \\
    SMSP(\textbf{ours})      & 88.55$\pm$0.26 & 90.24$\pm$0.35 \\
    \bottomrule
\end{tabular} }
\vspace{-2em}
\label{tab:random_mask}
\end{center}
\end{table}

\begin{figure}[tp]
     \centering
     \subfigure[Effect of number of similar tasks]{
         \centering
         \includegraphics[width=0.45\columnwidth]{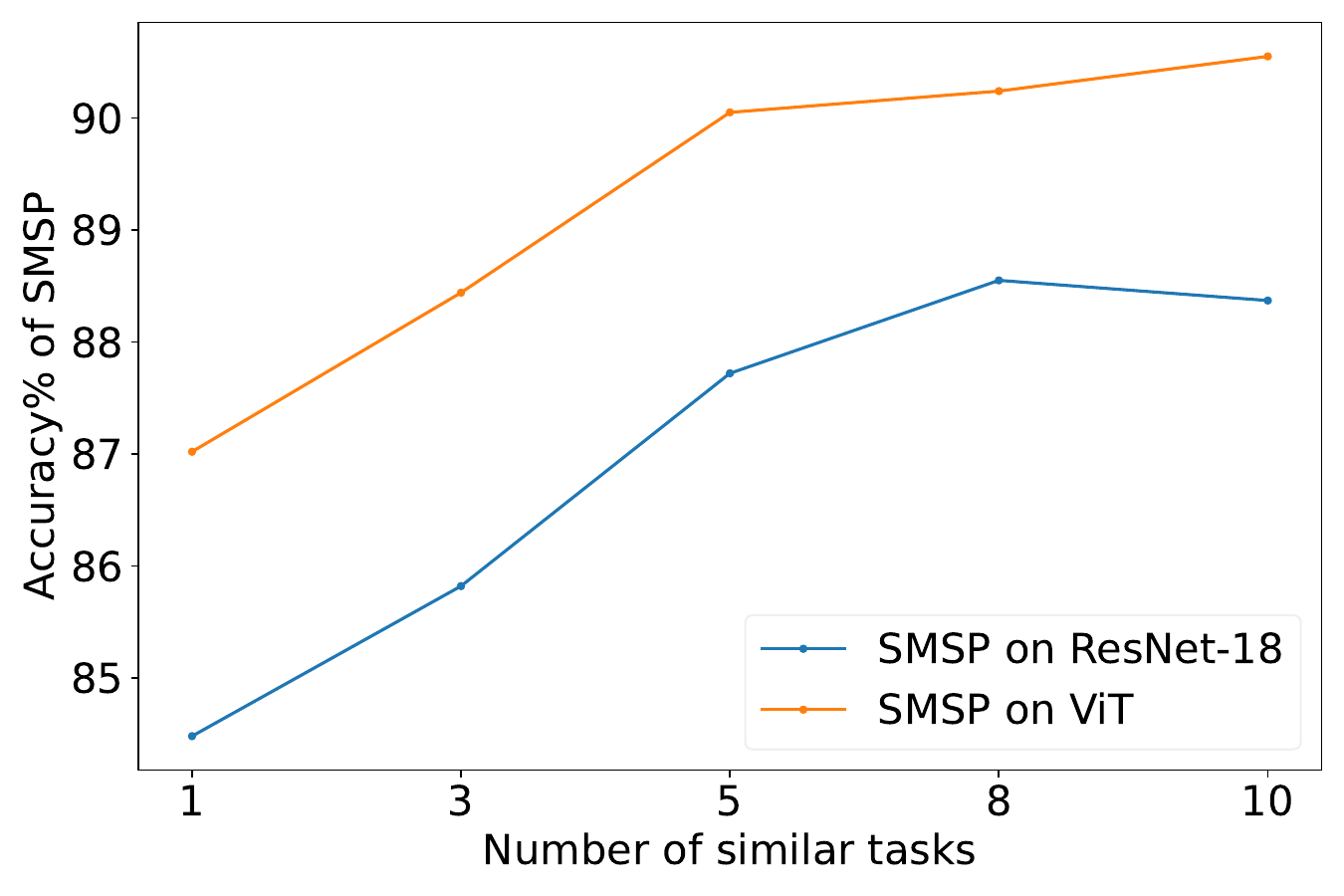}
     }
     \subfigure[Effect of task similarities]{
         \centering
         \includegraphics[width=0.45\columnwidth]{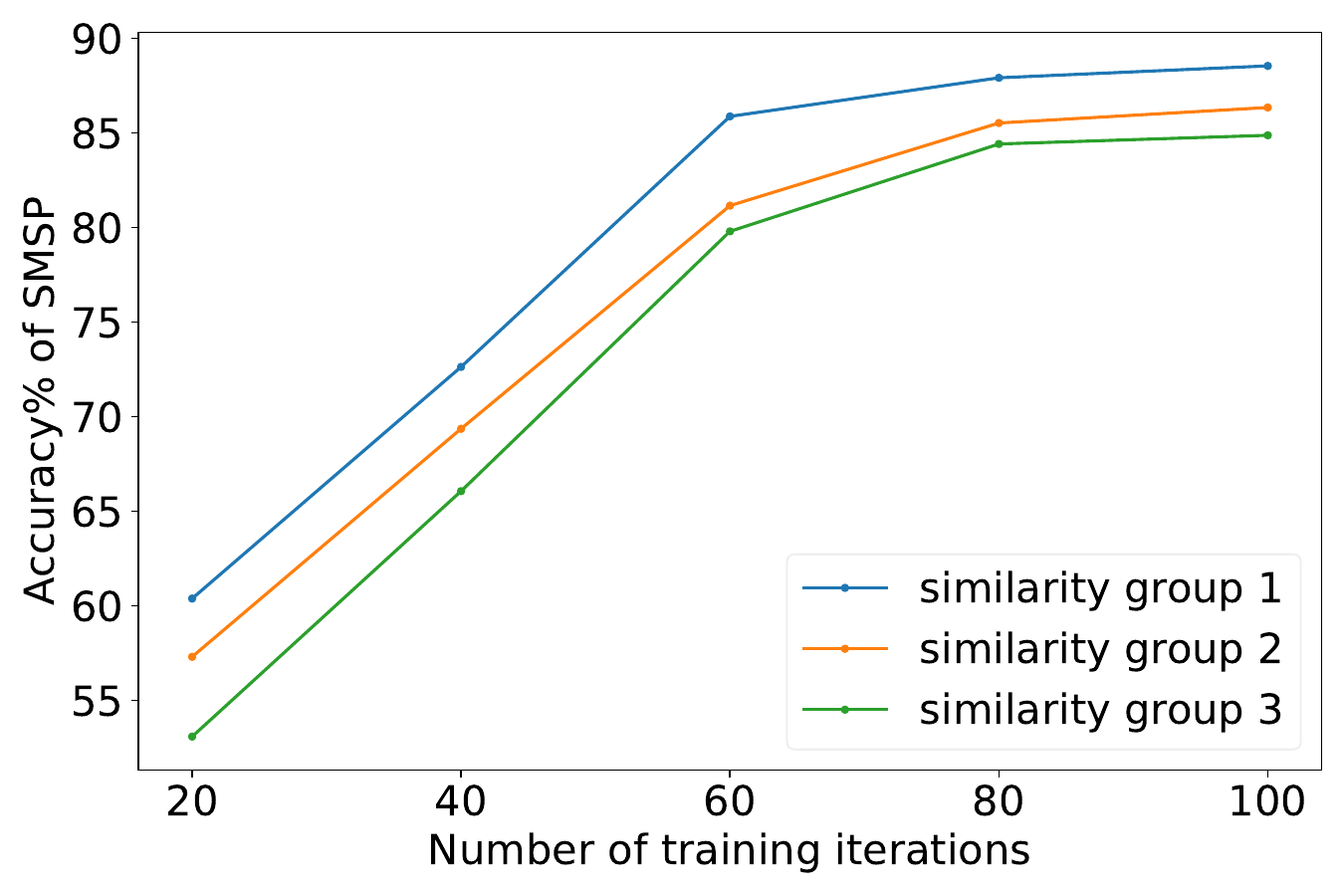}
     }
\caption{ (a) Results of applying SMSP to tasks of different pre-trained models over different numbers of similar tasks. More similar tasks (about 8) for each new task can lead to better performance for SMSP.
(b) Comparison between different similarity groups for SMSP with different training iterations. SMSP can achieve higher performance when tasks of higher similarities in the pool of pruned results are applied.
}
\label{fig:ablation_study}
\end{figure}

\textbf{Effect of the number of similar tasks.}
In plot (a) of Fig.\ref{fig:ablation_study}, we study the effect of the number of similar tasks for each new task. For both tasks from ResNet-18 and ViT, as the number of similar tasks increases, the performance of SMSP also improves.
This is because more pruned results of similar tasks can provide more task-specific knowledge for the new task. 
When the number $>8$, SMSP converges, which indicates that $8$ similar tasks for each task in SMSP are enough to create a high-quality mask.

\textbf{Effect of task similarities.}
In plot (b) of Fig.\ref{fig:ablation_study}, we compare the performance of SMSP when tasks with different similarities are used. The accuracy of using pruned models with higher similarities is always better than that of lower similarities, which implies that tasks with high similarities share more knowledge with new target tasks. This observation aligns with the findings presented in Section \ref{sec:emp_study}. The plot also illustrates that SMSP converges when the training iterations $>80$, indicating that only a limited number of training iterations will be enough for SMSP to build a promising pruned model.

\section{Conclusion}

In this paper, we propose a generic one-shot pruning method called SMSP to fast-adapt the pre-trained model to downstream tasks.
Based on the discovery that tasks with high similarities share more filters/nodes in their pruned models, given a new task, SMSP leverages the knowledge from the pruned models of its similar tasks to extract a sub-network from the pre-trained model. Then, a few training steps on the sub-network can reach a high-quality pruned model. Our experiments demonstrate that SMSP achieves SOTA results in terms of both accuracy and efficiency across various datasets and pre-trained models.

\bibliographystyle{splncs04}
\bibliography{mybibliography}

\end{document}